\documentclass[runningheads]{llncs}
\usepackage{graphicx}

\usepackage{tikz}
\usepackage{comment}
\usepackage{amsmath,amssymb} 
\usepackage{color}
\usepackage{tabu,multirow}
\usepackage{algorithm}
\usepackage{booktabs}
\usepackage{makecell}
\usepackage{algpseudocode}
\usepackage{subcaption}
\usepackage{wrapfig,lipsum,booktabs}
\def\etal{\textit{et~al.}}

\usepackage[accsupp]{axessibility}  

\begin{document}

\pagestyle{headings}
\mainmatter
\def\ECCVSubNumber{3127}  

\title{Towards Robust Face Recognition with Comprehensive Search} 


\titlerunning{Towards Robust Face Recognition with Comprehensive Search}
%
\author{Manyuan Zhang\inst{1,2} \and
 Guanglu Song\inst{2} \and
Yu Liu\inst{2}\textsuperscript{\dag}\and
Hongsheng Li\inst{1}}

\authorrunning{M. Zhang et al.}
%
\institute{Multimedia Laboratory, The Chinese University of Hong Kong \and
SenseTime Researsch \\
\email{zhangmanyuan@link.cuhk.edu.hk,songguanglu@sensetime.com,\\ liuyuisanai@gmail.com,hsli@ee.cuhk.edu.hk
}
\footnotetext{\textsuperscript{\dag} Corresponding author}
 }
\maketitle

\begin{abstract}

Data cleaning, architecture, and loss function design are important factors contributing to high-performance face recognition.
Previously, the research community tries to improve the performance of each single aspect but failed to present a unified solution on the joint search of the optimal designs for all three aspects.
In this paper, we for the first time identify that these aspects are tightly coupled to each other. Optimizing the design of each aspect actually greatly limits the performance and biases the algorithmic design.
Specifically, we find that the optimal model architecture or loss function is closely coupled with the data cleaning. 
To eliminate the bias of single-aspect research and provide an overall understanding of the face recognition model design,  we first carefully design the search space for each aspect, then a comprehensive search method is introduced to jointly search optimal data cleaning, architecture, and loss function design.
In our framework, we make the proposed comprehensive search as flexible as possible, by using an innovative reinforcement learning based approach.
Extensive experiments on million-level face recognition benchmarks demonstrate the effectiveness of our newly-designed search space for each aspect and the comprehensive search. We outperform expert algorithms developed for each single research track by large margins. 
 More importantly, we analyze the difference between our searched optimal design and the independent design of the single factors. We point out that strong models tend to optimize with more difficult training datasets and loss functions.
Our empirical study can provide guidance in future research towards more robust face recognition systems.
\keywords{Face Recognition, Comprehensive Search}
\end{abstract}

\section{Introduction}
\label{sec:intro}

Large-scale face recognition is a fundamental problem and of great practical value in   computer vision. It is challenging to learn robust feature representations from million-level datasets.
Recently, the vision community has rapidly improved the performance of  face recognition.
To a large extent, these advances have been driven by three aspects: large-scale noisy data cleaning strategies~\cite{guo2016ms,yi2014learning,cao2018vggface2,an2021partial,huang2008labeled}, margin-based loss function formulations~\cite{deng2019arcface,wang2018cosface,liu2017sphereface,wen2016discriminative},
and proper architecture designs~\cite{chen2018mobilefacenets,zhang2017polynet}.

\begin{figure}[t]
\begin{center}
\includegraphics[width=0.8\linewidth]{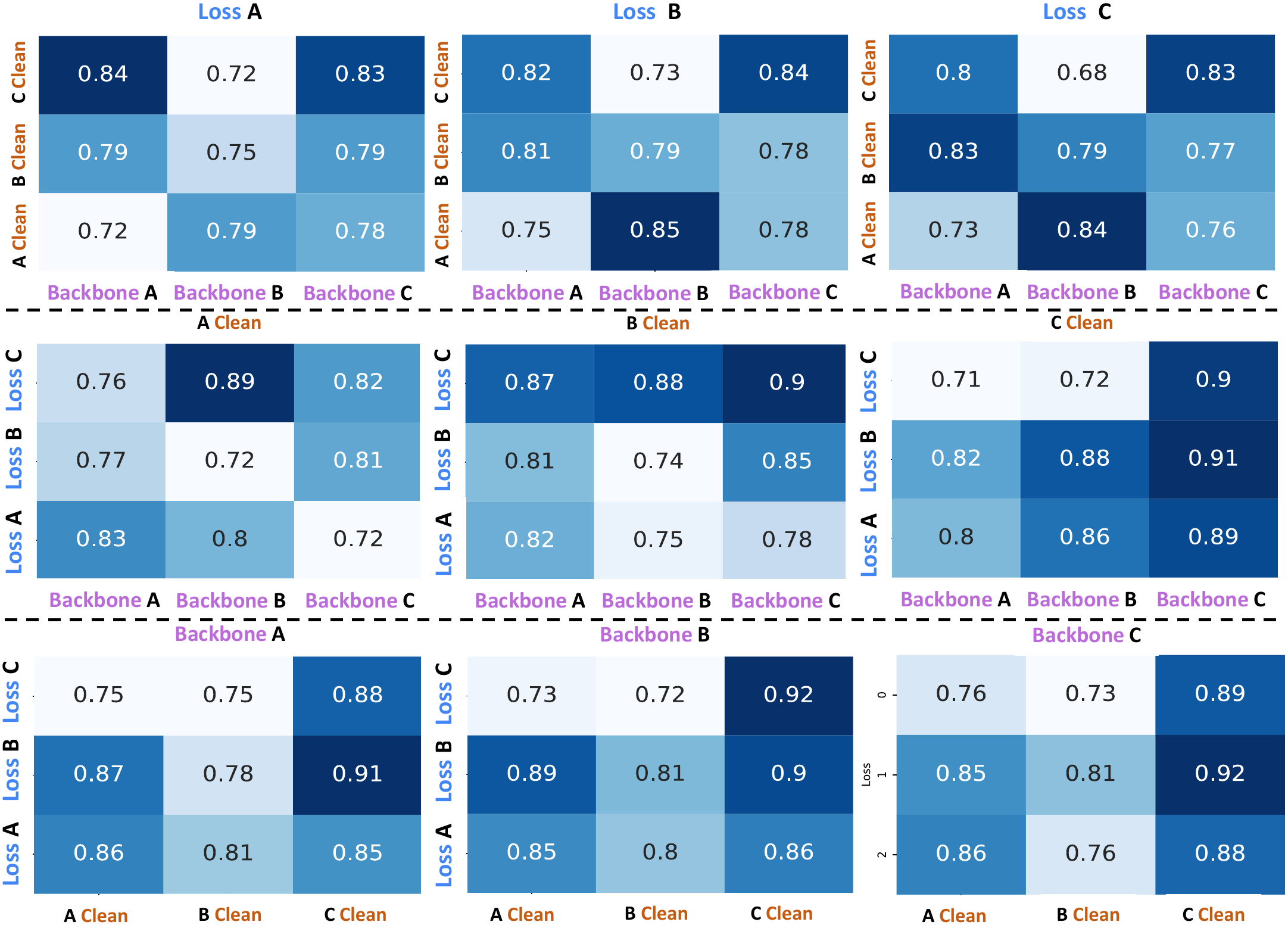}
\end{center}
   \caption{Given a fixed component (referred to the title of each table), we explore the optimal combinations between the other components (horizontal and vertical axes of each table).
    The optimal design of each component actually changes following the modifications of the other two, which suggests that it is sub-optimal to consider them separately. A, B, C represent different designs for each component. And the value is the validation accuracy.   }
\label{fig:pic1}
\end{figure}

Intuitively, large-scale datasets are important for face representation learning. 
Along with the development of various large-scale datasets~\cite{deng2019arcface,guo2016ms,wang2018devil,zhu2021webface260m,cao2018vggface2,an2021partial}, extensive works~\cite{wang2018devil,deng2020sub,wang2019co} tried to conduct automatic or semi-automatic data cleaning  to alleviate the influence of noisy data.
In addition, learning robust feature representation with margin-based loss functions~\cite{deng2019arcface,wang2018cosface,liu2017sphereface,zhang2019adacos,schroff2015facenet,sun2015deeply} can effectively enhance the representations. 
The  advanced architecture design also plays an import role for robust face recognition, including many common backbones ~\cite{he2016deep,szegedy2015going,zhang2018shufflenet,zhang2017polynet} and some specially designed backbones ~\cite{chen2018mobilefacenets,chang2021ressanet}.

However, there lacks a unified understanding of optimal designs of all the three aspects.
Previous methods all delved into one aspect while ignoring the other two, therefore neglecting the coupling between these aspects.  
For instance, many loss functions~\cite{schroff2015facenet,zhong2019unequal,hu2019noise} rely on hard sample mining to boost face recognition performance. However, this approach would be quite sensitive to noisy data. More specifically, many data cleaning methods~\cite{zhu2021webface260m,deng2019arcface} leverage the sample-to-class similarity scores to filter out outliers and merge similar classes. They pre-define a confidence threshold and samples with scores lower than the threshold are treated as the noise. Obviously, different thresholds would lead to different loss function designs.

To further evaluate this argument, we conduct extensive experiments to explore the coupling among these individual research aspects. We randomly select some combinations of different cleaning strategies,  loss function designs, and backbones. As shown in Figure~\ref{fig:pic1},  different combinations lead to significantly different performances.
The optimal design of each component actually changes following the modifications of the other two. Separate studies greatly limit the performance and would bias each aspect's algorithmic design.

To eliminate the bias of single-aspect research and provide an overall understanding of the face recognition models, we first propose new design space for each aspect, then deliberately design a comprehensive search method to jointly search data cleaning, architectures,  and loss function design. Undoubtedly, the proper design of search space  for each aspect will greatly contribute to the comprehensive search result.  For the data cleaning, we introduce the innovative sample discriminability-guided data cleaning search strategy to deal with the inter-and inter-class noise. We consider the relationship between a sample with its class centroid and the hardest-negative class to determine whether the sample is a noisy sample or not. It is worth mentioning that the search-based data cleaning approach can effectively alleviate the problem of tangled judgment on whether the edge samples are  hard positive or noisy samples. We clean the data simply based on whether the  searched cleanup result helps the training model form a more robust representation and whether it can cooperate well with the design of the other two aspects. For the loss function design formulation, we follow the successful margin-based loss function~\cite{deng2019arcface,wang2018cosface,wang2018additive}, but emphasize the importance of the scales of positive and negative samples, namely the scale-aware margin-based loss function design. As for the backbone design, we optimize the width and depth expansion ratio for the existing base, which have been shown to be the most important two factors in network design~\cite{tan2019efficientnet}. 

After carefully designing the search space for each aspect, we conduct the comprehensive search to jointly optimize them and explore their collaboration patterns.  Obviously, it is hard to simply integrate data cleaning, architectures,  and loss function design together.
The main barrier is how to effectively design and explore such a complex and huge search space.
To solve challenge, given the searchable formulation, we view the comprehensive search as a sequence prediction and generate their hyper-parameters as a sequence of tokens following the order of \texttt{\textbf{data}} \texttt{\textbf{cleaning}} $\xrightarrow{}$
\texttt{\textbf{loss}} \texttt{\textbf{function}} \texttt{\textbf{design}} $\xrightarrow{}$
\texttt{\textbf{architecture}} \texttt{\textbf{design}}.
Every prediction is produced by a softmax classifier and then fed into the next time step as input to influence the generation of the next hyper-parameter.
We adopt a Recurrent Neural Network (RNN) as the controller to explore the search space.
Once the controller RNN finishes generating the data clean strategy, an architecture, and a loss function, a joint solution based on these components is built and trained.
The parameters of RNN are then optimized to maximize the expected validation accuracy of the searched components.
The proposed comprehensive search effectively explores the learning of the coupling between different components
and enables a unified system.

Although some previous works utilize search-based methods for auto loss function~\cite{wang2020loss,li2019lfs} design, to our best knowledge, we are the first to jointly explore all three research fields and try to understand their entanglement.  And our proposed    scale-aware  margin-based loss function  can also surpass them substantially for the comparison in loss function design alone.  Extensive experiments show the superiority of our newly-proposed search space for each aspect, then the comprehensive search will further outperform former expert algorithms developed for each single research track by large margins. In summary, the main contributions of this paper can be summarized as follows:

\begin{itemize}
\item We identify an ignored problem of face recognition that the data cleaning strategy, loss function  design, and backbone architecture are coupled to each other. Separate studies on each factor greatly limit their performance.

\item We carefully design search space for each aspect and propose the  comprehensive search method to jointly explore data cleaning strategy, loss function formulation, and backbone architecture design. We introduce an innovative reinforcement learning based search framework to carry out the research and 
learns a unified system to achieve robust face recognition.

\item  We conduct extensive experiments on million-level face recognition tasks and evaluation on various benchmarks, including LFW, SLLFW, CALFW, CPLFW, IJB-B, and IJB-C.
It demonstrates the superiority of our search space design for each aspect and the huge performance gained from the integration together by the comprehensive search.

\end{itemize}

\section{Related work}

\subsection{Deep Face Recognition}

The proposal of large-scale noise-control datasets, strong backbone architectures, and well-designed loss functions have all greatly advanced the face recognition community. From the initial CASIA-Webface~\cite{yi2014learning} to the  recent  large-scale datasets Glint360K~\cite{an2021partial} and Webface260M~\cite{zhu2021webface260m}, these datasets have been proposed to greatly improve the accuracy of face recognition models. However,  these datasets rely on internet search engines and therefore contain a large percentage of noise. How to deal with these noises and learn face recognition with noisy labels has drawn much attention~\cite{zhong2019unequal,wu2018light,deng2020sub,zhu2021webface260m,liu2021switchable}.
Zhu~\cite{zhu2021webface260m} introduces the cleaning automatically by self-training (CAST) pipeline, which introduces a self-training process to auto clean inter-class and intra-class noise by an iteration manner and achieves good performance. For the model design, some architectures have been proposed to achieve efficient face recognition, such as PolyNet~\cite{zhang2017polynet} and MobileFaceNet~\cite{chen2018mobilefacenets}. For sophisticated face recognition loss function design, 
efficient margin-based softmax approaches~\cite{deng2019arcface,wang2018cosface,wang2018additive,liu2017sphereface,liu2019towards} were proposed by modifying the softmax loss function to achieve tighter intra-class compactness and more sparse inter-class separation, which achieve state-of-the-art performance. Since separate design on each component has achieved good performance, all the above methods ignore the relationship among data cleaning strategy, loss function formulation, and training backbone. Disentangled studies greatly limit performance. 

\subsection{Auto-ML for Face Recognition}

Auto-ML methods aim to automatically design suitable machine learning systems. Reinforcement learning guided search ~\cite{zoph2016neural,tan2019mnasnet} automate the design process and can easily outperform manually designed components in a given search space. 
Some former works use the searching methods to optimize the training loss function for face recognition. Li~\etal~\cite{li2019lfs} proposes the auto loss function search method from a hyper-parameter optimization perspective. Wang~\etal~\cite{wang2020loss} develop a unified formulation for the prevalent margin-based softmax loss. Then the random and reward-guided methods are designed to search for the best candidate. The above methods only focus on the loss function search. In this work, we consider all three aspects for face recognition. We first re-design the search space for each aspect, then jointly consider them and achieve excellent performance.

\section{Method}

To eliminate the bias of single-aspect research and provide an overall understanding of the face recognition model design, we propose the comprehensive search method to jointly search optimal data cleaning strategy, loss function formulation, and backbone architecture design for robust face recognition. In this section, we will introduce the details of our comprehensive search. The key components of comprehensive search include three aspects, \textit{i.e.} search space, search objective, and search algorithms. In the following, we will first introduce our newly-designed search space for each aspect,  then couple them together to form the comprehensive search space.   The search objective to evaluate the searched solution and the innovative reinforcement learning-based reward-guided comprehensive approach to identify the optimal combination design are introduced consequently.

\subsection{Comprehensive Search Space Design}

\subsubsection{Discriminabiliy-guided data cleaning search strategy.}

There are two kinds of data noise for face recognition datasets, the `outliers' and `label flips', corresponding to intra-class and inter-class noise respectively. An `outlier' noise means an image does not belong to any class of the datasets. A `label flip' noise refers to an image that is wrongly labeled with the incorrect class label. To clean the noise, we conduct inter-class filtering and intra-class merging. As for the intra-class noise cleaning, we filter the samples by their \emph{discriminability}~\cite{zhang2020discriminability}. The \emph{discriminability} is defined as: 

\begin{equation}
    \mathcal{D}_{i}=\frac{d i s t_{i p}}{\max \left\{ dist _{i n} \mid n \in[1, K], n \neq p\right\}},
\end{equation}

\noindent where $p$ is the  positive class label for sample $i$ and $n, n \in[1, K], n \neq p $ is the negative class label. $K$ is the number of class. $dist$ represents the feature similarity. We use the cosine similarity as the similarity indicator here. The
 \emph{discriminability} is the ratio between the feature’s similarity with the centroid of its class and the similarity from the hardest-negative class.  It can successfully distinguish the outliers and can be used to filter the samples whose \emph{discriminability} is lower than a  pre-defined confidence threshold $\tau_{intra}$.  As for the inter-class noise, we simply merge a pair of classes whose class center similarity is higher than the threshold $\tau_{inter}$.
The $\tau_{intra}$ and $\tau_{inter}$ are the two hyper-parameters that we need to optimize for the data cleaning strategy.  

One of the most difficult problems in past manual or semi-automatic data cleaning pipelines is how to classify edge samples as hard positive or noisy samples. Many loss functions~\cite{schroff2015facenet,sun2015deeply} rely on strategies such as hard sample mining to further boost performance.   However, edge samples are often hard to distinguish. By conducting the search-based data cleaning, we have eliminated this tangle. The result of data cleaning, \textit{i.e.} whether an edge sample is filtered or not, only depends on whether the cleanup result ultimately helps the training model to form a more robust feature representation, or whether it can cooperate well with the design of the other two aspects.

\subsubsection{Scale-aware margin-based loss function design.}

Margin-based softmax loss functions ~\cite{deng2019arcface,wang2018additive,wang2018cosface,liu2017sphereface} have been proposed in recent years to enhance the feature discrimination for face recognition. In summary, they can be defined in a uniform formulation. Suppose  $\boldsymbol{w}_{k} \in \mathbb{R}^{d}$ is the $k$-th class's weight $(k \in\{1,2, \ldots, K\})$ and $K$ is the number of classes. $\boldsymbol{x} \in \mathbb{R}^{d}$ denotes the face feature. The  formulation for the margin-based loss function can be written as follows:

\begin{equation}
    \mathcal{L}=-\log \frac{e^{s_{p} f\left(m, \theta_{w_{y}, x}\right)}}{e^{s_{p} f\left(m, \theta_{w_{y}, x}\right)}+\sum_{k \neq y}^{K} e^{s_{n} \cos \left(\theta_{w_{k}, x}\right)}},
\end{equation}

\noindent where $\theta_{\boldsymbol{w}_{k}, \boldsymbol{x}}$ is the angle between $\boldsymbol{w}_{k}$ and $\boldsymbol{x} $. Suppose  $\cos \left(\theta_{\boldsymbol{w}_{k}, \boldsymbol{x}}\right)=\boldsymbol{w}_{k}^{T} \boldsymbol{x}$ is the cosine similarity of the $k$-th class weight and  feature, $f\left(m, \theta_{\boldsymbol{w}_{y}, \boldsymbol{x}}\right) \leq \cos \left(\theta_{\boldsymbol{w}_{y}, \boldsymbol{x}}\right)$ is the carefully designed margin function and  can be summarized into the combined version 
$f\left(m, \theta_{\boldsymbol{w}_{y}, \boldsymbol{x}}\right)=\cos \left(m_{1} \theta_{\boldsymbol{w}_{y}, \boldsymbol{x}}+m_{2}\right)-m_{3}$. $m_{1}$ and $m_{2}$ are the multiplicative and additive angular margin, and $m_{3}$ is the additive cosine margin. Previous work~\cite{deng2019arcface,wang2018cosface,wang2018additive} has mainly focused on the margin optimization, but ignored the importance of tuning the positive sample scale  $s_{p}$  and the negative sample scale $s_{n}$, which  control the optimization difficulty of positive and negative sample. We experimentally prove that face recognition accuracy can be further hugely improved by optimizating the $s_{p}$ and $s_{n}$ jointly. Overall, $ m_{1},m_{2},m_{3}, s_{p} $ and $s_{n} $ 
are the hyper-parameters we need to search for the loss function.

\begin{figure*}[t]
\begin{center}
\includegraphics[width=\linewidth]{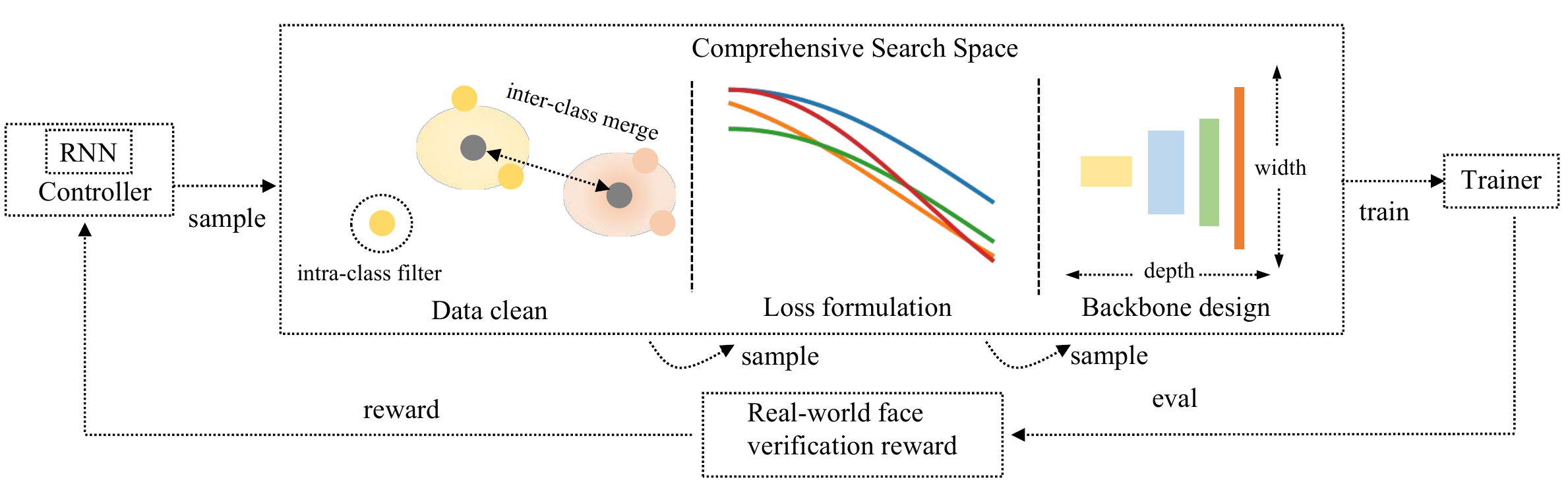}
\end{center}
  \caption{The overall pipeline for our innovative reinforcement learning-based  comprehensive search for robust face recognition. We design it for the sample-eval-update loop.  Firstly, the RL agent  samples a batch of  combinations that contains the hyper-parameters for  data cleaning strategy, loss function formulation, and backbone architecture design. For each sampled combination, we train it on the target face recognition task to get its accuracy and reward.  After that, the agent will be updated by maximizing the expected reward until converged. }
\label{fig:pipeline}
\end{figure*}

\begin{algorithm}[t]

\renewcommand{\algorithmicrequire}{\textbf{Input:}}
\renewcommand{\algorithmicensure}{\textbf{Output:}}
\caption{Comprehensive Search}\label{alg:cma}
\begin{algorithmic}
\Require Training set $\mathcal{S}_{t}=\left\{\left(\boldsymbol{x}_{i}, y_{i}\right)\right\}_{i=1}^{n} $, validation set $\mathcal{S}_{v}$, total searching epochs $T$, agent policy $\pi_{\theta}$ and its initialized parameter $\theta_{0}$.

\For{$t=1$ to $T$ do}

\State Sample $B$ combinations of hyper-parameters $\boldsymbol{C}_{1}, \ldots \boldsymbol{C}_{B}$ for data cleaning  strategy $\tau_{intra}$, $\tau_{inter}$, loss function  formulation $ m_{1},m_{2},m_{3}, s_{p} $, $s_{n}$ and backbone architecture design $\mathcal{W}$ and $\mathcal{D} $ via RNN sequencely.

\State Train the combinations $\boldsymbol{C}_{1}, \ldots \boldsymbol{C}_{B}$ for one epoch separately on the training set $\mathcal{S}_{t}=\left\{\left(\boldsymbol{x}_{i}, y_{i}\right)\right\}_{i=1}^{n} $ with the sampled hyper-parameters.

\State Evaluate the trained combinations $\boldsymbol{C}_{1}, \ldots \boldsymbol{C}_{B}$ on the validation set $\mathcal{S}_{v}$ to get reward $\mathrm{R}(\boldsymbol{C}_{1}),\ldots \mathrm{R}(\boldsymbol{C}_{B}) $ via Eq.~\ref{eq2}.

\State Update $\theta_{t}$ by $\max\mathbb{E}_{P} \mathrm{R}(\boldsymbol{C})$ 
\State Update $\pi_{\theta_{t+1}}=\pi_{\theta_{t}}$

\EndFor

\Ensure Final policy  $\pi_{\theta}$

\end{algorithmic}
\end{algorithm}
\subsubsection{Effective depth  and width search for the backbone.}

Following ~\cite{tan2019efficientnet}, the width expansion ratio $\mathcal{W}$ and the depth expansion ratio $\mathcal{D} $ are  the two most important factors for neural network architecture design, so we utilize those two factors to search 
backbone architecture for face recognition. The base model we selected is modified MobileNet~\cite{howard2017MobileNets}, which has been proved successful for face recognition and its search cost is affordable.

\subsection{Comprehensive Search Objective and  Algorithm}

Joint search with data cleaning strategy,  loss function design, and architecture is not easy to perform due to the complex and huge search space.
To enable fast and flexible comprehensive search, 
we view the comprehensive search as a sequence prediction and generate their hyper-parameters as a sequence of tokens following the order of \texttt{\textbf{data}} \texttt{\textbf{cleaning}} $\xrightarrow{}$
\texttt{\textbf{loss}} \texttt{\textbf{function}} \texttt{\textbf{design}} $\xrightarrow{}$
\texttt{\textbf{architecture}} \texttt{\textbf{design}}. Every prediction of each step is produced by a softmax classifier and then fed into the next time step as input to influence the generation of the next hyper-parameter. We adopt the Recurrent Neural Network (RNN) as the controller to generate the search  parameters. Once the controller has completed the generation  of  the data clean strategy, the loss function, and the architecture, a joint solution based on these components is built and trained. The parameters of RNN are optimized to maximize the expected validation accuracy of the searched components. The overall searching process can be optimized with the sample-eval-update loop as shown in Alg.~\ref{alg:cma} and Figure~\ref{fig:pipeline}.

To solve the real-world face recognition problem, we test all candidate joint solutions and use the performance as the search criterion.
Another object that needs to be considered is the computational cost, since we search for the depth  expansion ratio and width expansion ratio for the backbone, we expect the new backbone to have a similar computational budget with the original backbone. To this end, we use a weighted product method to approximate Pareto optimal solutions. The final reward function can be formulated as:

\begin{equation}
\label{eq2}
    \mathrm{R}(\boldsymbol{C})=\mathrm{ACC}(\boldsymbol{C}) \times\left[\frac{\mathrm{COST}(\boldsymbol{C})}{\mathrm{TAR}}\right]^{\alpha} 
\end{equation}

\noindent where $\boldsymbol{C}$ is the sampled combination of designs of the three aspects, $\operatorname{ACC}(\boldsymbol{C})$ is the accuracy on real-world face verification task of the sampled combination, $\operatorname{COST}(\boldsymbol{C})$ is the computation cost (FLOPs) of the combination, $\operatorname{TAR}$  is the target computation cost which we set it to the original backbone, and $\alpha$ is the weight factor. 

 Following~\cite{tan2019mnasnet}, we use Proximal Policy Optimization (PPO)~\cite{schulman2017proximal} to optimize the RL agent to find Pareto optimal solutions for our  comprehensive search problem. For each sampled combination in the search space, we map it to a list of tokens, which are determined by a sequence of action $a_{1: T}$ from the RL agent with policy $\pi_{\theta}$. The overall objective is to maximize the expected reward:
\begin{equation}
\label{eq4}
    \mathbb{J}=\mathbb{E}_{\left.P_{(} a_{1: T} ; m\right)} \mathrm{R}(\boldsymbol{C})
\end{equation}

For each sample $\boldsymbol{C}$, we train it on the target task to get its accuracy $\operatorname{ACC}(\boldsymbol{C})$ and its computation cost $\operatorname{COST}(\boldsymbol{C})$. We then calculate the reward $\mathrm{R}(\boldsymbol{C})$ using Eq.~\ref{eq2}. After that, the agent with policy $\pi_{\theta}$ will be updated by maximizing the expected reward in Eq.~\ref{eq4}
until the parameter $\theta$ converges. Then we will re-train the top combinations   $\boldsymbol{C}$ with the highest  $\mathrm{R}(\boldsymbol{C})$ by full train and select the best combination as our searched results. 

\section{Experiment}

\subsection{Datasets}

\subsubsection{Training Data.} To conduct the comprehensive search for robust face recognition, we use the MS1MV2~\cite{deng2019arcface} as our training set, which contains  5.8 million face images from 85K identities. To prove the good generation ability of our comprehensive search, we also experiment on the recently introduced large-scale dataset Glint360K~\cite{an2021partial}, which contains 17 million images
of 360K individuals. Note that the MS1MV2 and Glint360K are the two largest publicly available face recognition datasets.

\subsubsection{Test Data.} During the  sample-eval-update loop, we evaluate our combination on the re-organised MegaFace~\cite{kemelmacher2016megaface} verification benchmark. After retraining, we test the final model at most popular face benchmarks including LFW~\cite{huang2008labeled}, CALFW~\cite{zheng2017cross}, SLLFW~\cite{deng2017fine}, CPLFW~\cite{zheng2018cross}, IJB-B~\cite{whitelam2017iarpa} and IJB-C~\cite{maze2018iarpa}. 
We follow the unrestricted with labelled outside data protocol~\cite{huang2008labeled}.

\subsection{Implementation Details}

\subsubsection{Data Processing.} All the faces in the training images are detected by RetinaFace~\cite{deng2019retinaface}.  Alignment by five landmarks is conducted and the face is cropped to 112$\times$112. Images are normalized by subtracting 127.5 and dividing by 128. For the data cleaning strategy, we use the ResNet-50~\cite{he2016deep} model trained on MS1MV2 to extract  feature to conduct intra-class and inter-class cleaning.

\subsubsection{Searching.} During the searching process, each sampled solution is trained on the training set and evaluated on the validation set. The weighted accuracy of TAR@FAR at $10^{-3}$, $10^{-4}$, $10^{-5}$ by 0.5, 0.25 and 0.25  are used as the $\mathrm{ACC}$.  After that, we re-train the top reward solution. We choose the top 20 solutions under the computation budget and fully train them on the target task and then report the final performance on  popular face benchmarks.  A recurrent neural network (RNN) is utilized as the controller to generate the parameter combination. We use Adam optimizer with  learning rate of $5\times10^{-4}$ and momentum 0.9 to update the controller. For the sampled solution's training, we only train each solution for one epoch. The search process needs around 1,000  samples  to converge and costs around 37 GPU days (NVIDIA A100, FP16 training).

\subsubsection{Retraining.} After choosing  the top 20 combinations, we fully retrain them. We use SGD optimizer with weight decay $5\times10^{-4}$ and momentum 0.9.  We train models on 8 NVIDIA GPUs with batch size 1024 for 100K iterations. The initial learning rate is set to 0.1 and decays by 0.1 at iterations 40K, 60K, and 80K for MS1MV2. As for Glint360K, we train for batch size 1024 with 150k iterations totally and learning rate decay at 60k, 90k, and 13k.

\begin{table}[t]
\centering
\caption{Verification performance (\%) of different search space combinations on the LFW, SLLFW, CALFW, CPLFW, SLLFW, IJB-B and IJB-C benchmarks. The last row represents the baseline performance, which combines the best previous hand-crafted design for each track. The training set is MS1MV2.}
\def\arraystretch{1.2}
\resizebox{0.85\textwidth}{!}{%
\begin{tabular}{ccc|cccc|cc|cc}
\toprule
\multicolumn{3}{c|}{Search Space} & \multirow{2}{*}{LFW} & \multirow{2}{*}{SLLFW} & \multirow{2}{*}{CALFW} & \multirow{2}{*}{CPLFW} & \multicolumn{2}{c|}{IJB-B} & \multicolumn{2}{c}{IJB-C} \\ \cline{1-3} \cline{8-11} 
Data      & Loss      & Backbone  &                      &                        &                        &                        & $10^{-5}$         & $10^{-4}$       & $10^{-5}$        & $10^{-4}$         \\ \hline \hline
$\checkmark$ & $\checkmark$ & $\checkmark$ & 99.55                & \textbf{98.78}                  & \textbf{94.75 }                 & \textbf{84.80}                   & \textbf{85.66}       & \textbf{91.46 }       & \textbf{90.45}       & \textbf{93.58}       \\
$\times$     & $\checkmark$ & $\checkmark$ & \textbf{99.58}                & 98.43                  & 94.15                  & 84.33                  & 83.67       & 91.20        & 89.55       & 93.24       \\ \hline
$\checkmark$ & $\times$     & $\times$     & 99.43                & 98.15                  & 94.00                  & 82.48                   & 82.35       & 90.72        & 88.47       & 92.84       \\
$\times$     & $\checkmark$ & $\times$     & 99.52                & 98.27                  & 94.03                   & 84.30                   & 83.26       & 90.80        & 88.88       & 92.91       \\
$\times$     & $\times$     & $\checkmark$ & 99.35                & 98.22                  & 94.23                  & 82.78                  & 80.62       & 90.40        & 87.92       & 92.70       \\ \hline
$\times$     & $\times$     & $\times$     & 99.43                & 98.12                  & 94.03                  & 82.30                  & 79.71       & 90.44       & 87.29       & 92.61       \\ \bottomrule
\end{tabular}}%

\label{tab:tab1}
\end{table}

\subsection{Results}\newlength{\oldintextsep}
\setlength{\oldintextsep}{\intextsep}

\setlength\intextsep{0pt}
\begin{wraptable}{r}{0pt}
\centering
\def\arraystretch{1.3}
\resizebox{0.4\linewidth}{!}{
\begin{tabular}{ccccccccc}
\toprule
\multicolumn{9}{c}{Best combination}                                               \\ \hline
$\tau_{intra}$ & $\tau_{inter}$ & $m_{1}$  & $m_{2}$   & $m_{3}$   & $s_{p}$ & $s_{n}$ & $\mathcal{D} $ & $\mathcal{W}$ \\ \hline \hline
0.3         & 0.62         & 1.15 & 0.22 & 0    & 40    & 48       & 1.47  & 0.84   \\
0.22         & 0.76         & 1.0 & 0.32  & 0    & 40    & 40       & 1.22  & 0.91   \\
0.26          & 0.62         & 1.2 & 0.36 & 0 & 32    & 32       & 1.47  & 0.84  \\ \bottomrule
\end{tabular}
}
\caption{The  best combinations searched with our comprehensive search on MS1MV2 dataset.}
\label{tab:best_combinations}
\end{wraptable}

We test our comprehensive search method on widely-used face verification benchmarks including LFW, SLLFW, CALFW, CPLFW, IJB-B, and IJB-C. The results are shown in Table~\ref{tab:tab1}. The bold numbers in each column represent the best results.  In Table~\ref{tab:tab1}, the last line represents the baseline results without any search-based design component. For the baseline, the training dataset is semi-automatically cleaned by~\cite{deng2019arcface}, the loss function is finely designed margin-based loss function ArcFace~\cite{deng2019arcface}, and the backbone architecture is modified MobileNet~\cite{howard2017MobileNets} that is specially designed for face recognition.  
From the results, we can see that our comprehensive search outperforms the baseline by a large margin on all test benchmarks, \textit{i.e.} 0.97\% performance gain for IJB-C at 1e-4 and 3.16\% at 1e-5. The huge improvement demonstrates the great potential of comprehensive search compared to manually design components separately. What's more, the search for each component also boosts the performance greatly, which shows the excellent and flexible design of our newly-designed search space for each component and the limitation of hand-crafted design. From the result of row 1 and rows 3-5 in Table~\ref{tab:tab1}, we can also observe that searching for each component separately may not achieve the best performance. If we comprehensively search for all three aspects jointly, we can further boost the performance significantly. Comparing rows 1 and  2 in Table~\ref{tab:tab1}, the addition of search for data cleaning strategy improves the search result dramatically, which has been ignored in previous searching methods~\cite{li2019lfs,wang2020loss}.
We show some combinations of top verification accuracy we searched on MS1MV2 in Table~\ref{tab:best_combinations}. The results show a very different design preference of searching from the previous manual design for each component separately.

\begin{figure}[t]
\begin{center}
\includegraphics[width=0.6\linewidth]{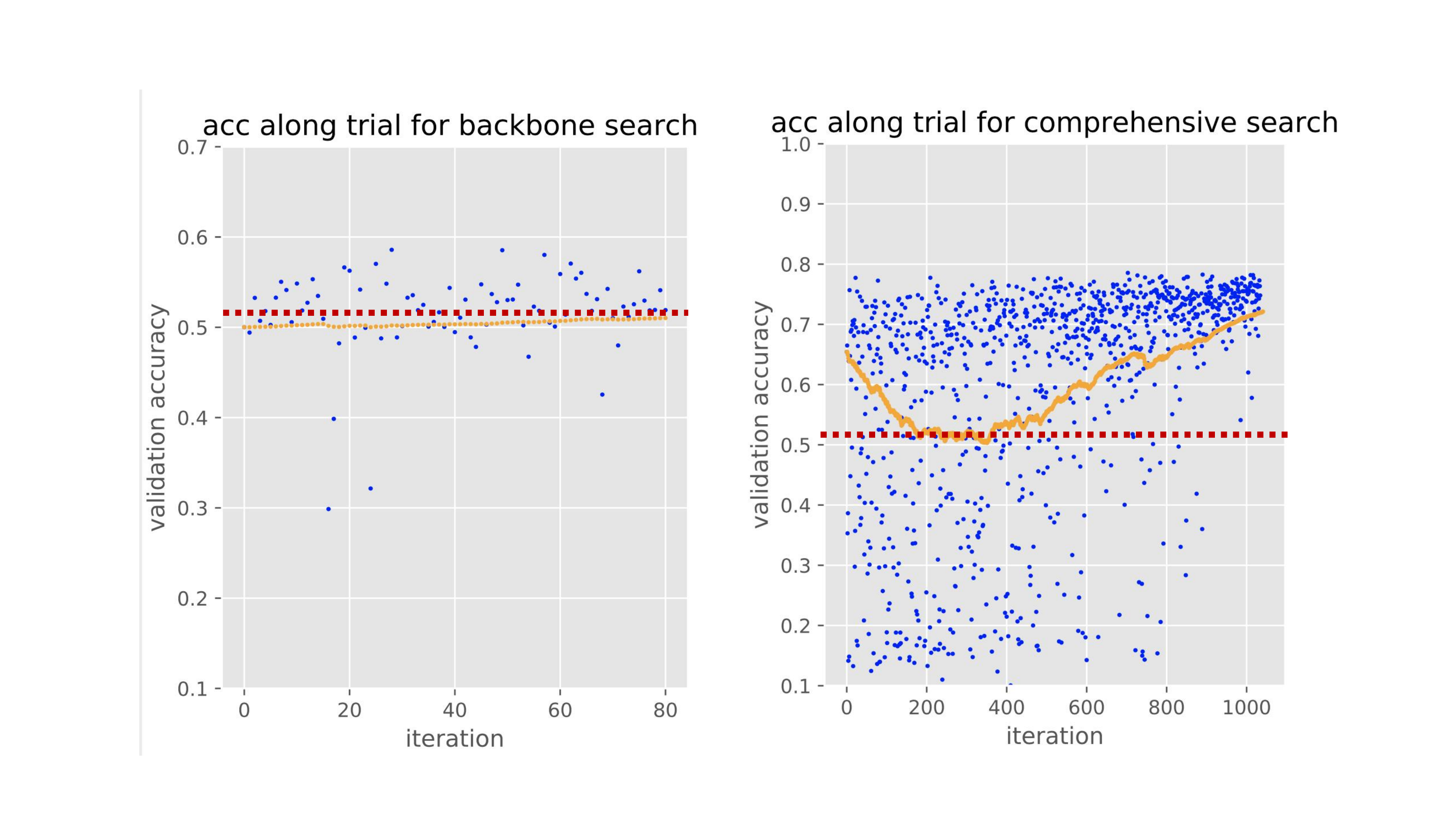}
\end{center}
   \caption{ Visualization of the  accuracy on validation set along the search process. \textbf{Left}: The  accuracy for searching backbone architecture only. 
  \textbf{Right}: The  accuracy for comprehensive search. 
 The yellow line represents the moving average  of all sampled combinations, and the red dot line indicates the baseline. }
\label{fig:searching_processing}

\end{figure}

\subsection{Searching Process}

We visualize the searching process in Figure~\ref{fig:searching_processing}. The yellow line represents the moving average of the accuracy of the sampled solution. From the figure, we can observe that  the  validation accuracy has been improved gradually through the searching, which suggests that the agent has learned a robust policy that can sample high-quality hyper-parameter combinations. We also compare the searching process for comprehensive search and backbone architecture search in Figure~\ref{fig:searching_processing}. From the figure, we can also 
see that by enhancing with  flexible data cleaning strategy and loss function formulation searching, the  validation accuracy of sampled combination  has been improved greatly, which confirms the validity of our comprehensive search. In addition, all the rewards of comprehensive and backbone search outperform the baseline combination (represented by the red dot line), which shows the effectiveness of searching.

\subsection{Ablation study}

\subsubsection{Effect of \emph{discriminabiliy}-guided data cleaning strategy.} For the data cleaning strategy, we propose the \emph{discriminabiliy}-guided data intra-class filtering and inter-class merging according to  class center similarity to determine `outlier` and `label flip` samples respectively. From the results of rows 3 and 6 in Table~\ref{tab:tab1},  we can observe that only  searching for data cleaning strategy can outperform baseline whose dataset is semi-automatically cleaned hugely, and achieve performance gains of 1.18\% at $10^{-5}$ on IJB-C and 2.64\%  at $10^{-5}$ on IJB-B benchmark.  
Note that the result is customized, as we have stated before, the best candidate for cleaning hyper-parameters is highly related to the loss function formulation and backbone architecture design.

\subsubsection{Effect of cleaning feature extraction model.}
\setlength{\oldintextsep}{\intextsep}
\setlength\intextsep{0pt}
\begin{wraptable}{r}{0pt}
\centering
\resizebox{0.25\linewidth}{!}{
\renewcommand{\arraystretch}{1.3}
\begin{tabular}{c|cc}
\toprule
\multirow{2}{*}{Train Dataset} & \multicolumn{2}{c}{IJB-C}          \\ \cline{2-3} 
                              & \multicolumn{1}{c|}{$10^{-5}$}  & $10^{-4}$  \\ \hline \hline                       MS1MV2 & 88.80 & 92.88 \\ 
Glint360K                       & \multicolumn{1}{c|}{\textbf{88.84}} & \textbf{92.92} \\ \bottomrule
\end{tabular}
}
\caption{Results of different feature extracted model.}
\label{tab:tab_feature_extraction}
\end{wraptable} In the cleaning process, we rely on the feature extracted by a  pre-trained model  to calculate \emph{discriminabiliy} and inter-class similarity. Intuitively, the more discriminate features would lead to better cleaning accuracy. We conduct an ablation study to study the influence. We use the ResNet-50 backbone trained on MS1MV2 and Glint360K to extract features and the results are shown in Table~\ref{tab:tab_feature_extraction}.  The better features would lead to better cleaning accuracy, but the impact is quite limited.

\subsubsection{Effect of   scale-aware  margin-based loss function.}  For the loss function design, we formulate it as the scale-aware margin-based loss function.  The results of rows  4 and 6 in Table~\ref{tab:tab1} demonstrate that  the search of loss function   can improve the performance significantly compared to the widely used hand-crafted design loss ArcFace,  \textit{i.e.} 1.59\% gain at IJB-C $10^{-5}$. Furthermore, compared to the other two components data and backbone, the separate searching of loss achieves the largest improvement, which demonstrates that sophisticated loss function design is crucial for learning discriminate feature representation.

\subsubsection{Effect of  depth and width search.} For backbone architecture design, we only search for the width expansion ratio $\mathcal{W}$ and the depth expansion ratio $\mathcal{D}$. We constrain the new backbone to share the same FLOPs with the base model.   From Table~\ref{tab:tab1} rows 5 and 6, the simple width and depth search can still boost performance, which indicates the limitation of hand-crafted backbone design.

\begin{table}[t]
\caption{ Verification performance (\%) of different search space combinations on LFW, SLLFW, CALFW, CPLFW, SLLFW, IJB-B and IJB-C benchmarks. The last row represents the baseline performance, which combines the best previous hand-crafted design for each track. The training set is Glint360K. }
\centering
\def\arraystretch{1.2}
\resizebox{0.9\textwidth}{!}{
\begin{tabular}{ccc|cccc|cc|cc}

\hline
\multicolumn{3}{c|}{Search Space} & \multirow{2}{*}{LFW} & \multirow{2}{*}{SLLFW} & \multirow{2}{*}{CALFW} & \multirow{2}{*}{CPLFW} & \multicolumn{2}{c|}{IJB-B} & \multicolumn{2}{c}{IJB-C} \\ \cline{1-3} \cline{8-11} 
Data      & Loss      & Backbone  &                      &                        &                        &                        & $10^{-5}$& $10^{-4} $& $10^{-5}$        & $10^{-4}$        \\ \hline \hline
$\checkmark$ & $\checkmark$ & $\checkmark$ & \textbf{99.63  }              & \textbf{98.75 }                 & \textbf{94.28}                  & \textbf{86.20 }                 & \textbf{86.06}        & \textbf{92.55}       & \textbf{90.06  }     & \textbf{94.36 }      \\
$\times$     & $\times$     & $\checkmark$ & 99.38                & 98.11                  & 94.07                  & 85.10                   & 80.96        & 91.13       & 86.68       & 93.25       \\ \hline
$\times$     & $\times$     & $\times$     & 99.34                & 98.05                  & 93.83                  & 84.35                  & 79.83        & 90.92       & 86.33       & 93.16       \\ \hline
\end{tabular}}
\label{tab:tab3}
\end{table}

\subsection{Search on Glint360K}

To show the good generalizability of our comprehensive search, we also conduct  experiments on the newly introduced large-scale dataset Glint360K~\cite{an2021partial}. The results are shown in Table~\ref{tab:tab3}. From the table, we can see that the comprehensive search improves the performance significantly on all face benchmarks, which demonstrates  good generalizability. Our solution improves 3.73\% at $10^{-5}$ for IJB-C  and  6.23\% at $10^{-5}$ for IJB-B  respectively. The main reason is that our proposed comprehensive search  can effectively capture the intrinsic connections among the different aspects and find the best combinations. Moreover, the best combinations in Glint360K of the three aspects are different from the results searched on MS1MV2, which validates our observation that the three aspects are coupled to each other.

\begin{table}[t]
\begin{center}
\makebox[0pt][c]{\parbox{\textwidth}{%
\begin{minipage}[t]{0.5\hsize}
    \caption{Verification performance (\%) of  dataset transferability experiment on IJB-B and IJB-C.}
    \resizebox{\linewidth}{!}{
    \renewcommand{\arraystretch}{1.4}
        \setlength\tabcolsep{3.5pt}
    \begin{tabular}{c|c|cc|cc}
\toprule
\multirow{2}{*}{Search on} & \multirow{2}{*}{Train on} & \multicolumn{2}{c|}{IJB-B} & \multicolumn{2}{c}{IJB-C} \\ \cline{3-6} 
                          &                           & $10^{-5}$         & $10^{-4}$        & $10^{-5}$        & $10^{-4}$        \\ \hline \hline
Glint360K                  & MS1MV2                 & \textbf{84.26}        & \textbf{91.12}       & \textbf{89.41}       & \textbf{93.18}       \\ 
MS1MV2                    & MS1MV2                 & 85.66        & 91.46       & 90.45       & 93.58       \\ 
-                          & MS1MV2                 & 79.71        & 90.44       & 87.29       & 92.61       \\ \bottomrule
\end{tabular}}%
    \label{tab:tab4}%
\end{minipage}
\hfill
\begin{minipage}[t]{0.5\hsize}

    \caption{ Verification performance (\%) of comprehensive search for ResNet50  on  IJB-B and IJB-C benchmarks with MS1MV2. The last row represents the baseline performance.}
    
    \resizebox{\linewidth}{!}{
    \renewcommand{\arraystretch}{1.21}
    \setlength\tabcolsep{10pt}
    \begin{tabular}{cc|cc|cc}
    \toprule

\multicolumn{2}{c|}{Transfer} & \multicolumn{2}{c|}{IJB-B} & \multicolumn{2}{c}{IJB-C} \\ \hline
Data          & Loss          & $10^{-5}$ & $10^{-4}$        & $10^{-5}$        & $10^{-4}$        \\ \hline \hline
$\checkmark$& $\checkmark$          &       \textbf{90.11}       &      \textbf{94.55}       &      \textbf{94.11}       &   \textbf{96.04}          \\ \hline
$\times$          & $\times$          & 89.11        & 94.24       & 93.68       & 95.76       \\ \bottomrule
\end{tabular}}%
    \label{tab:tab5}%
\end{minipage}
}}
\end{center}
\end{table}

\subsection{Transferability}

Since the three aspects are highly related 
and the best combination would be changed for different  training datasets, we  explore the transferability of our comprehensive search in this subsection. We use the best combination searched on Glint360K to train  models on MS1MV2.  Results are shown in Table~\ref{tab:tab4}. The results show that even the transfer learning reduces performance gains, we still achieve significant performance improvements compared to the baseline.
The result also  suggests that it is better to search and test with the same dataset and  the best candidate of one component would be changed along with the other two.

\subsection{Search with larger backbone}

In the above experiments, we have performed the search based on the modified MobileNet whose FLOPs is around 0.33G. In this section, we  perform a joint search based on ResNet50 whose FLOPs is around 6G. We search for data cleaning strategy and loss function, and the results are shown in Table~\ref{tab:tab5}. The performance improvement verifies the generalization of our joint search.

\subsection{Discussion}

In order to explore the intrinsic relationship among these three aspects, we  analyse the results of the search to find the best combination pattern. To facilitate the analysis, we defined the difficulty of the data and the difficulty of the loss function separately. For data cleaning, a lower $\tau_{intra}$ as well as a higher $\tau_{inter}$ implies greater training difficulty. For loss function, stricter margins  and larger $S_{n}/S_{p}$  imply more difficult optimizations. They are defined as follows,
\begin{equation}
\begin{split}
       & \rm Difficulty_{data}=1-\tau_{intra}+\tau_{inter} \\
    & \rm Difficulty_{loss}=\frac{s_{n}*(m_{1}-1+m_{2}+m_{3})}{s_{p}}
\end{split}
\end{equation}

\begin{figure}[t]
     \centering
     \begin{subfigure}[b]{0.30\textwidth}
         \centering
         \includegraphics[width=\textwidth]{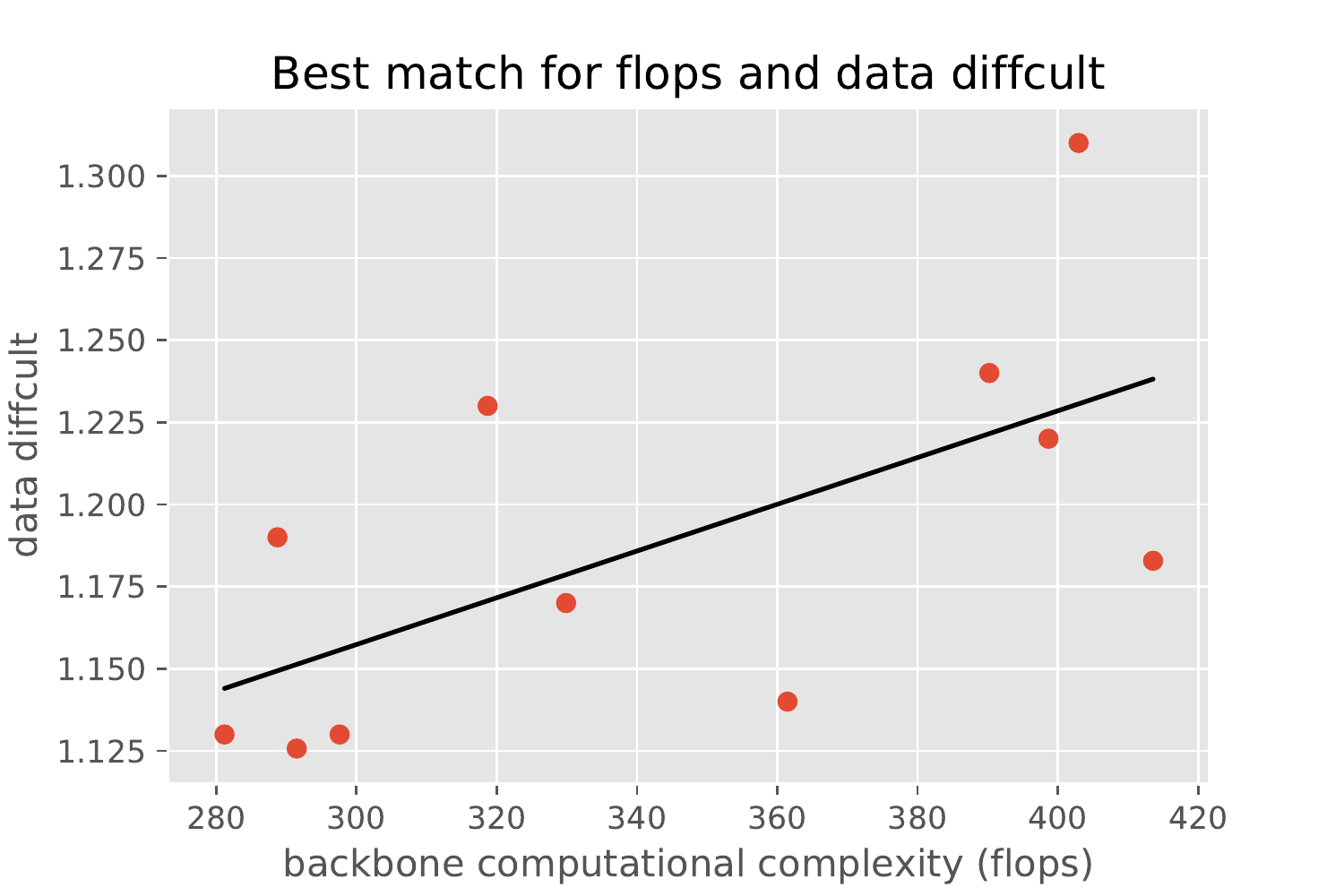}
         \caption{}
         \label{fig:sub1}
     \end{subfigure}
     \hfill
     \begin{subfigure}[b]{0.30\textwidth}
         \centering
         \includegraphics[width=\textwidth]{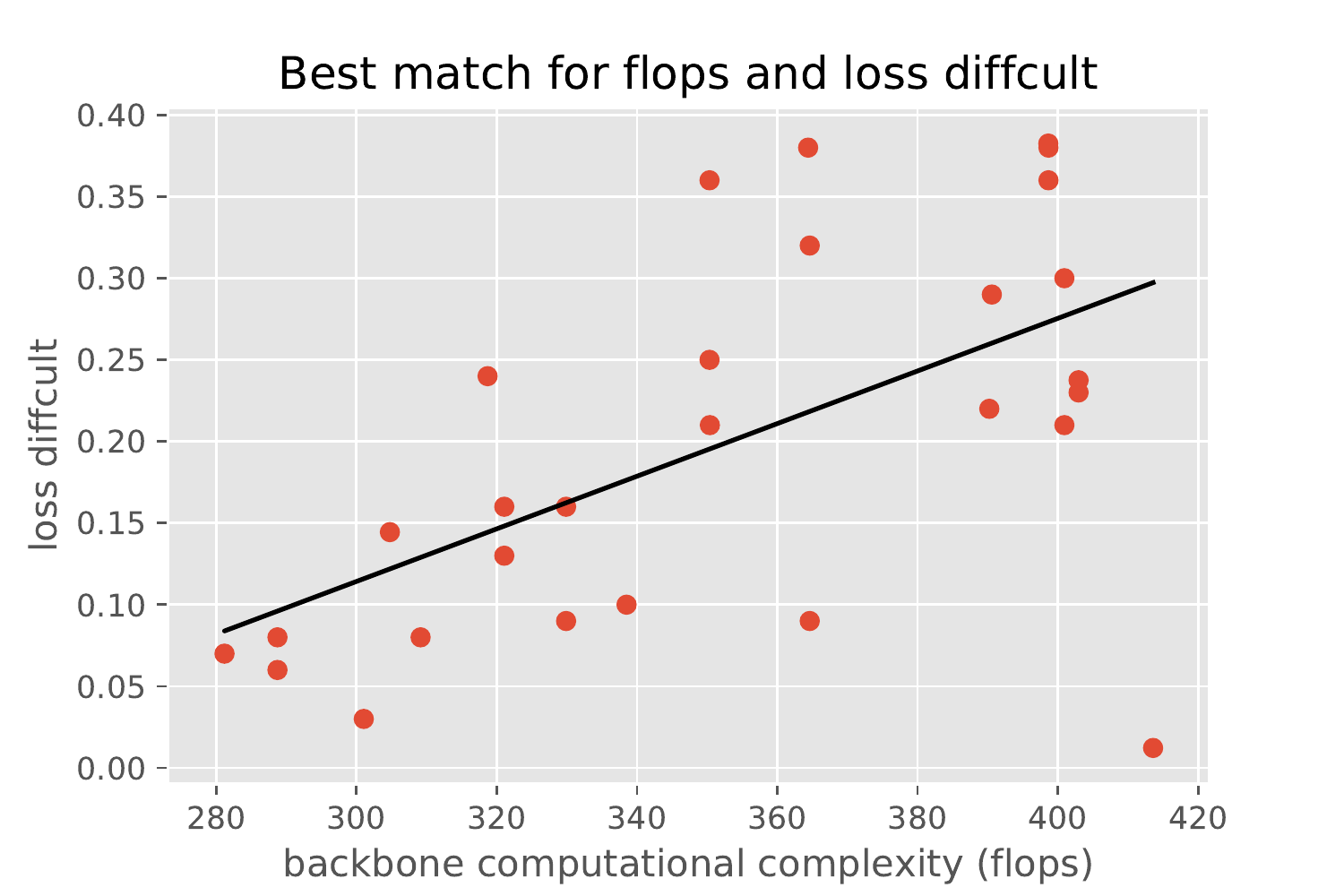}
         \caption{}
         \label{fig:sub2}
     \end{subfigure}
     \hfill
     \begin{subfigure}[b]{0.30\textwidth}
         \centering
         \includegraphics[width=\textwidth]{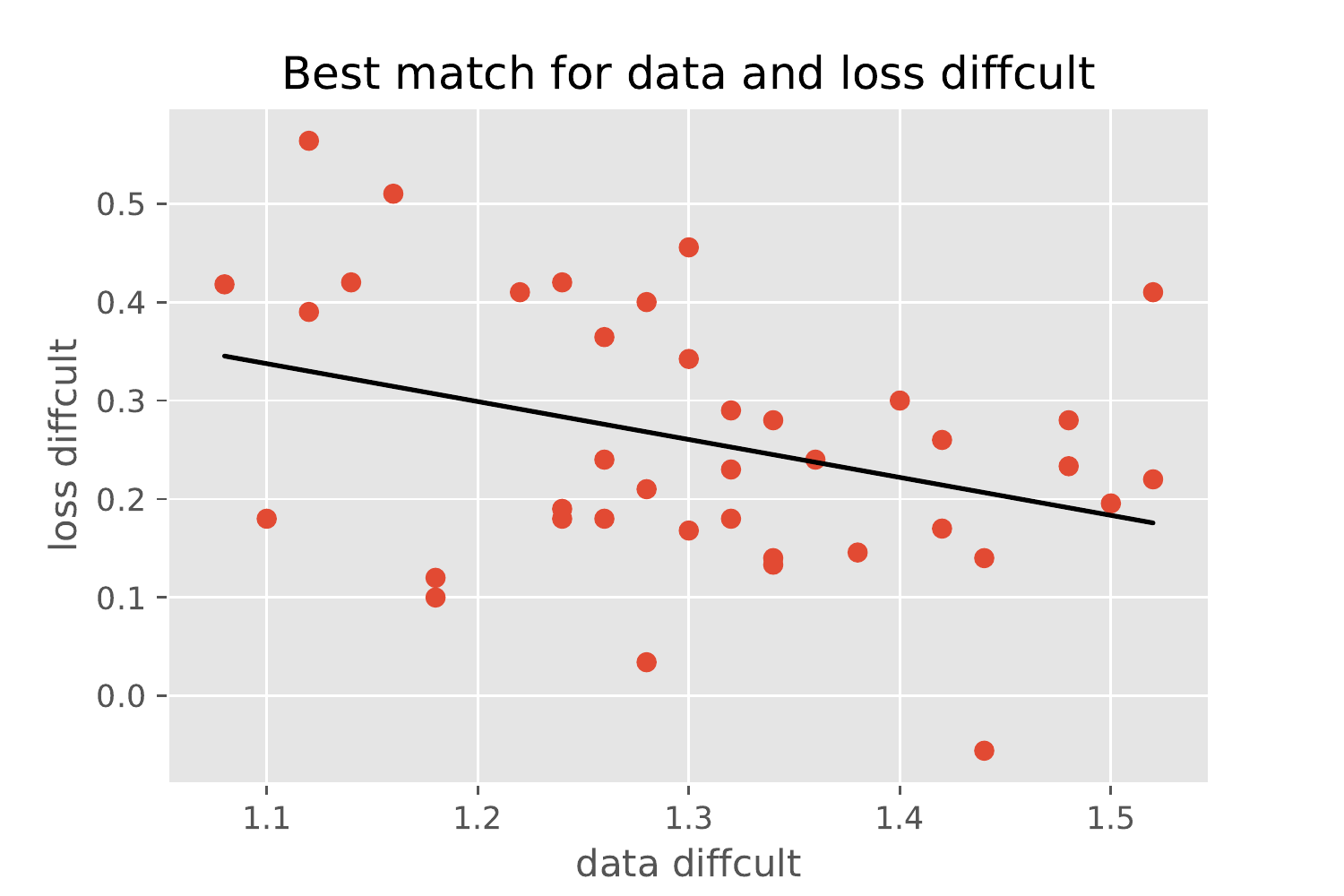}
         \caption{}
         \label{fig:sub3}
     \end{subfigure}
        \caption{Visualization of best matches among  data difficult, loss difficult  and backbone  computational complexity (FLOPs).}
        \label{fig:relation}
\end{figure}

We visualize the relationship between data difficulty, loss difficulty, and backbone architecture  in  Figure~\ref{fig:relation}. In Figure~(\ref{fig:sub1}) and~(\ref{fig:sub2}), larger models (higher FLOPs) tend to train with greater data difficulty and loss difficulty. A larger model implies a stronger fitting ability and is, therefore, able to handle more complex optimization problems. So it can  learn more from difficult samples and strict optimization objectives. For the best matching of loss function and  data cleaning, as Figure~(\ref{fig:sub3}) suggests, models tend to choose easier optimization loss functions under severe data. These findings provide a fresh perspective for the design of  face  recognition pipeline.

\section{Conclusion}

In this work, we explore the relationship among the data cleaning strategy, loss function formulation, and backbone architecture design for robust face recognition. Previously, people optimize them separately but fail to present a unified understanding of integrating them. We provide a fresh perspective, coupling them and optimizing jointly.  We  propose  an innovative reinforcement learning-based comprehensive search  for the best combination. Extensive experiments have proven the excellence of our method for robust face recognition. 

\section*{Acknowledgement}

Hongsheng Li is also a Principal Investigator of Centre for Perceptual and Interactive Intelligence Limited (CPII). This work is supported in part by CPII, in part by the General Research Fund through the Research Grants Council of Hong Kong under Grants (Nos. 14204021, 14207319), in part by CUHK Strategic Fund.

\bibliographystyle{splncs04}
\bibliography{egbib}
\end{document}